\documentclass[runningheads]{llncs}

\usepackage[T1]{fontenc}
\usepackage{graphicx}
\usepackage{url}
\usepackage{hyperref}
\usepackage{booktabs, array, amsmath}
\begin{document}

\title{Cognitive Agent Compilation for Explicit Problem Solver Modeling}

\titlerunning{Cognitive Agent Compilation}


\author{
Hyeongdon Moon\inst{1}\orcidID{0000-0001-5759-2017} \and
Carolyn Rosé\inst{1}\orcidID{0000-0003-1128-5155} \and
John Stamper\inst{1}\orcidID{0000-0002-2291-1468}
}
\institute{
Carnegie Mellon University, Pittsburgh PA 15213, USA \\
\email{\{donim,cprose,jstamper\}@andrew.cmu.edu}
}
\authorrunning{H. Moon et al.}

\maketitle

\begin{abstract}
Large language models (LLMs) are widely used for tutoring, feedback generation, and content creation, but their broad pretraining makes them hard to constrain and poor substitutes for controllable learners. Educational systems often require inspectable and editable knowledge states: educators want to know what a system assumes the learner knows, and learners benefit when the system can justify actions in terms of explicit skills, misconceptions, and strategies. Inspired by cognitive architectures, we propose Cognitive Agent Compilation (CAC), a framework that uses a strong teacher LLM to compile problem-solving knowledge into an explicit target agent. CAC separates (i) knowledge representation, (ii) problem-solving policy, and (iii) verification and update rules, with the goal of making bounded problem solving more inspectable and editable in educational settings. We present an early proof of concept implemented with Small Language Models that surfaces key design trade-offs, particularly between explicit control and scalable generalization, and positions CAC as an initial step toward bounded-knowledge AI for educational applications.
\end{abstract}


\section{Introduction}

Large Language Models (LLMs) have been rapidly integrated into educational environments, with both students and teachers utilizing them for a variety of tasks. This trend is equally prevalent in Artificial Intelligence in Education (AIED) research, where numerous studies propose employing LLMs as teachers or proxy students to alleviate resource bottlenecks. However, a common issue across these attempts is the discrepancy between LLMs and actual human cognition. Substantial evidence suggests that even Chain of Thought or reasoning models operate fundamentally differently from human thought processes~\cite{mahowald2024dissociating}. Most notably, LLMs exhibit a significant weakness in their inability to ``suppress knowledge generation,'' that is, to simulate a state of not knowing. This inherent limitation renders raw LLMs unsuitable as faithful proxies for student learners.

LLMs have already demonstrated performance approaching or exceeding that of human experts across many problem-solving domains. Yet ``knowing'' a concept and ``explaining it so that another can understand'' are distinct cognitive abilities. Even human subject-matter experts frequently struggle to teach effectively without proper pedagogical training~\cite{nathan2001expert}. Consequently, it is questionable whether LLMs, which have absorbed vast amounts of unconstrained information and thus possess extensive knowledge without pedagogical intuition, can effectively fulfill their potential in educational applications. The critical question remains: how can an LLM effectively scaffold knowledge acquisition for a human who lacks that very knowledge?

To address this, we propose the Cognitive Agent Compilation (CAC) framework, which compiles bounded and inspectable problem-solving behavior into an explicit target agent. The Teacher LLM generates explicit, executable artifacts such as knowledge items, applicability conditions, and update rules that define the agent's behavior. If a cognitive agent is constructed using explicit, deterministic operations rather than relying entirely on opaque LLM internals, its behavior can be attributed more directly to declared knowledge and control structure. By restricting the Teacher LLM's role to generating and revising explainable knowledge artifacts, CAC aims to expose which knowledge assumptions are necessary for success. The explanatory value of the resulting system depends on how much of the final behavior is carried by explicit structure rather than latent prior knowledge. For this reason, we treat CAC not as a validated cognitive replica, but as a bounded and inspectable problem-solving model that may support explanatory learner modeling research~\cite{DBLP:journals/bjet/RoseMLK19}.

This perspective suggests several possible applications in the learning sciences, including more inspectable and editable problem-solving models for educational settings. To explore this direction, we present a proof of concept implemented with Small Language Models and use it to surface key technical and conceptual challenges that arise when compiling educationally meaningful problem-solving structure from pretrained language models.

\section{Related Works}
\subsection{Explanatory learner models and transparency}
Explanatory learner models emphasize that the purpose of a learner model is not merely prediction, but supporting explanations, diagnosis, and interventions; machine learning alone often under-specifies the representational commitments education needs~\cite{DBLP:journals/bjet/RoseMLK19}. Open learner models extend this logic by inviting learners into the model, enabling reflection and negotiation over the system's beliefs~\cite{BullKay2007SMILI}. More broadly, XAI in education identifies distinct explainability requirements tied to pedagogy, accountability, and stakeholder roles~\cite{Khosravi2022XAIED}. These lines motivate CAC's primary goal: an agent with a declared, editable knowledge state and explanations grounded in that state.

\subsection{Learner modeling and cognitive model refinement}
Knowledge Tracing models procedural knowledge acquisition through explicit latent variables updated over interaction logs~\cite{CorbettAnderson1994KT}. Deep knowledge tracing improved prediction by learning implicit representations of knowledge states~\cite{Piech2015DKT}, but at the cost of interpretability. In the cognitive modeling lineage, Learning Factors Analysis (LFA) provides a method for evaluating and improving cognitive models by proposing alternative decompositions into knowledge components (KCs) and testing their predictive adequacy~\cite{CenKoedingerJunker2006LFA}. Automated Student Model Improvement pushes this further by iteratively refining student models with data~\cite{Koedinger2012ASMI}. CAC uses these ideas as both design anchors (KCs as explicit units) and potential evaluation targets (does an explicit decomposition pay off?).

\subsection{Students as agents and tutor authoring}
SimStudent and apprentice learner models treat a learner as an agent that can be taught, enabling tutor authoring and analysis of instructional strategies~\cite{Matsuda2011SimStudent}. Example-tracing tutors show how explicit traces of correct behavior can structure tutoring decisions~\cite{Aleven2009ExampleTracing}, while domain-general tutor authoring with apprentice learner models aims to reduce domain-specific engineering overhead~\cite{MacLellanKoedinger2022ApprenticeAuthoring}. CAC inherits the same commitment: educational systems benefit from explicit, manipulable representations of problem solving.

\subsection{Cognitive Architectures}

Cognitive Architectures represent a research lineage spanning over four decades, aiming to reproduce human cognitive processes through explainable algorithms. Representative models such as ACT-R, SOAR, and EPIC have been designed for various cognitive modeling purposes. Among these, ACT-R has served as the theoretical foundation for major learning science methodologies, including Bayesian Knowledge Tracing~\cite{CorbettAnderson1994KT}, Performance Factor Analysis~\cite{pavlik2009performance}, and Additive Factor Models~\cite{CenKoedingerJunker2006LFA}.

Because individual cognitive units must be explicitly defined as executable code, traditional Cognitive Architectures allow for highly precise modeling but suffer from limited scalability due to the domain expert labor required for their construction. Even before the advent of LLMs, there were attempts to integrate natural language processing into these systems, as seen in NL-Soar~\cite{lehman2006gentle}. More recently, the generalization performance of LLMs has catalyzed initiatives to fuse them with foundational cognitive architectures, exemplified by CoALA~\cite{sumers2023cognitive}. The present study inherits and explores this synergistic research trajectory.

\section{The CAC framework}
\subsection{Core idea}
CAC is a pipeline for converting a strong model's problem-solving capability into a target agent that operates over an explicit knowledge representation, minimizing opaque reasoning where possible. The goal is not merely compression, but explicitness: the deployed agent should expose its knowledge state (what it knows, what it does not know, what misconceptions it holds), and its behavior should be attributable to that state.

\begin{figure}[ht]
    \centering
    \includegraphics[width=0.85\linewidth, trim=0cm 8.5cm 15.5cm 0cm, clip]{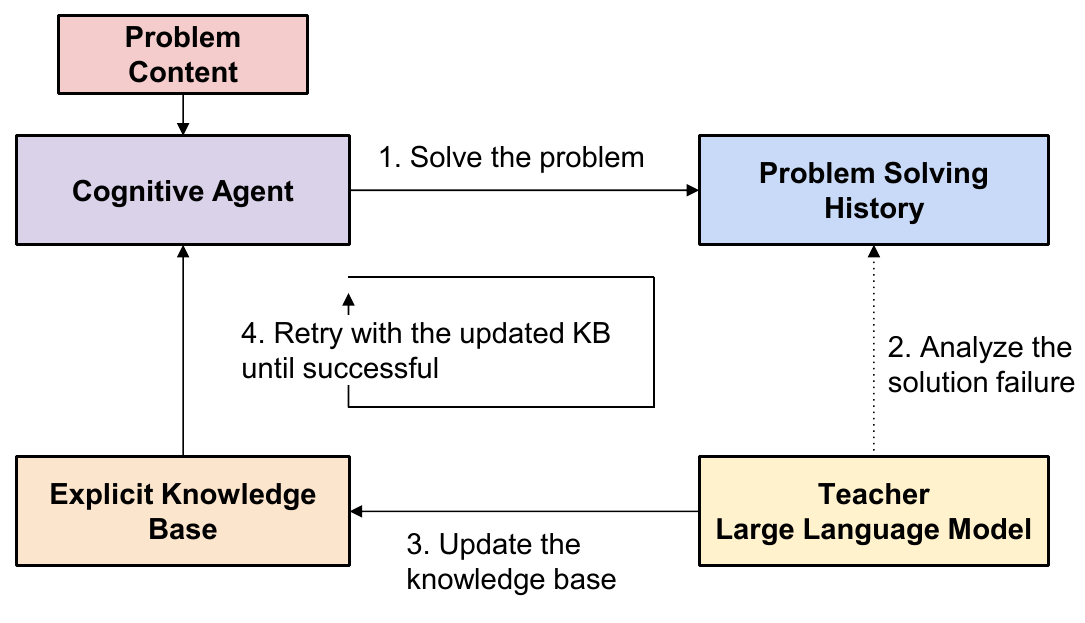}
    \caption{Overview of the Cognitive Agent Compilation (CAC) framework. The framework operates through a failure-driven learning cycle where the Cognitive Agent attempts a problem , generating a Problem Solving History. The Teacher Large Language Model analyzes these traces for failures or suboptimal paths and distills corrective knowledge into an explicit Knowledge Base (KB) , prompting the agent to retry until successful}
    \label{fig:main}
\end{figure}
\subsection{Inputs and outputs}

\subsubsection{Inputs}
\begin{itemize}
 \item \textbf{Problem Content}: Task specifications, domain constraints, and specific problem instances targeting candidate KCs and misconceptions.
 \item \textbf{Teacher Large Language Model}: The strong model equipped with tools (solvers, evaluators, symbolic executors) acting as the diagnostic and compilation engine.
 \item \textbf{Target-agent specification}: The structural definition of the \textbf{Cognitive Agent} and its \textbf{Explicit Knowledge Base} (e.g., KC graph, rule base, permitted representation types).
\end{itemize}

\subsubsection{Outputs}
\begin{itemize}
 \item A refined \textbf{Cognitive Agent} with an updated, human-readable \textbf{Explicit Knowledge Base} and a verified \textbf{Problem Solving History} that documents the knowledge acquisition process.
\end{itemize}

\subsection{The CAC loop}
The CAC framework operationalizes a failure-driven learning cycle, iterating through four specific steps as illustrated in the system diagram. Fig. \ref{fig:main} visualizes the sequence of the CAC loop.

\subsubsection{Step 1: Solve the problem.} 

The \textbf{Cognitive Agent} receives the \textbf{Problem Content} and attempts to solve it by strictly relying on its current \textbf{Explicit Knowledge Base}. This attempt generates a detailed \textbf{Problem Solving History}, recording intermediate reasoning steps, applied procedural rules, and retrieved knowledge chunks.

\subsubsection{Step 2: Analyze the solution failure.}

The \textbf{Teacher Large Language Model} acts as an evaluator and diagnostician. It inspects the \textbf{Problem Solving History} against process constraints and final correctness. If a failure or suboptimal path is detected, the Teacher analyzes the trace to identify the specific missing knowledge components (KCs), misconception triggers, or misapplied strategies responsible for the error.

\subsubsection{Step 3: Update the knowledge base.} 

Based on the diagnostic analysis, the Teacher formulates the necessary corrective knowledge and injects it back into the \textbf{Explicit Knowledge Base}. Unlike generic self-refinement~\cite{Madaan2023SelfRefine}, these updates are compiled into explicit, inspectable artifacts such as new rule bases, program sketches, or restructured retrieval items, rather than being left implicit in latent weights.

\subsubsection{Step 4: Retry with the updated KB until successful.}

The \textbf{Cognitive Agent} re-attempts the problem. Because its underlying \textbf{Explicit Knowledge Base} has been fundamentally modified in Step 3, the agent's reasoning trajectory alters accordingly. This verification and update cycle repeats until both final answers and intermediate process constraints are fully satisfied, confirming that the agent solves the task using the updated knowledge base.

\subsection{Designing the Knowledge Base}
Defining the representational schema of the knowledge stored in the Knowledge Base and designing how the Cognitive Agent depends on it is the most critical component of the CAC framework. For instance, if the Cognitive Agent is based on ACT-R, where all production rules and declarative memory can be defined as explicit Lisp code snippets, the knowledge could be formalized as Lisp code pieces adhering to valid ACT-R syntax. Consequently, the Teacher LLM would require a profound understanding of ACT-R's operational rules to generate these structures.

The core principle of designing this knowledge schema is twofold: first, the Teacher LLM must be able to predict the behavioral shifts resulting from the addition or removal of specific knowledge elements; second, it must be able to correct the Cognitive Agent's problem-solving trajectory solely by modifying the Knowledge Base, without intervening in any other cognitive components. While classical cognitive architectures can articulate this mechanism with relative clarity, they carry the risk of overwhelming the Teacher LLM with an excessive amount of context required for exact code generation. To address this limitation, an example of designing a more scalable, natural language-based Knowledge Base is introduced in Section \ref{sec:5}.

\section{Design Objectives and Intended Applications}
\label{sec:goals}

The CAC framework was designed with several educational goals in mind, but the current paper does not validate them against student traces. The discussion below should therefore be read as a set of design targets and evaluation criteria rather than established capabilities.

\subsection{Student simulation under limited knowledge}
\label{sec:simulation}

A primary design goal of CAC is to support the estimation of a student's problem-solving trajectory under a restricted knowledge pool by selectively removing specific knowledge from the Knowledge Base. If such interventions consistently change problem resolvability in predictable ways, they would offer one possible route toward a more explicit treatment of the Knowledge Tracing question: ``Can problem resolvability be predicted given a student's specific knowledge state?''

Provided the prior probability distribution of each knowledge element is known and the compiled agent is validated against student behavior, such a setup could offer one possible route toward predicting problem resolvability from symbolic knowledge states in ITS interaction logs.
Existing Knowledge Tracing methodologies often fail to model Knowledge Components (KCs) with sufficient granularity to handle ``multiple solutions to a single problem'' or ``multi-step problems.'' In contrast, CAC is intended to make the assumptions about required KCs explicit, thereby supporting more inspectable intervention design.

\subsection{Identifying well-defined Knowledge Components}
\label{sec:goodkc}

A second design target is to provide semantic intuition regarding what constitutes a ``well-defined'' KC. If a specific KC extracted through CAC can be applied across various problem types addressing the same underlying knowledge, it suggests broadly defined knowledge rather than problem-specific trivia.

Related work on evaluating knowledge dependency argues that a question is educationally useful insofar as solving it depends on the targeted knowledge rather than on broad background knowledge alone~\cite{moon2022evaluating}. CAC may therefore complement KC identification methodologies such as Learning Curve Analysis, which heavily rely on empirical data and are inherently susceptible to noise.

\subsection{Toward declarative-first individual difference modeling}
\label{sec:individual}

Sections~\ref{sec:simulation} and~\ref{sec:goodkc} both operate at the level of declarative memory: identifying what a learner knows or does not know in terms of explicit, retrievable knowledge elements. This focus is deliberate. Cognitive architecture research has long recognized that repeated execution of a declarative knowledge chain leads to progressive automatization, described as chunking in Soar~\cite{lehman2006gentle} and production compilation in ACT-R~\cite{anderson2004integrated}. Yet no practical algorithm exists to reverse-engineer these grouping processes at the individual level. We suggest that beginning with the declarative layer offers an efficient starting point from which to later investigate how that chain compiles into procedural skill through individual-specific processes. CAC is designed as this first step.

\section{Proof of concept: Implementation with SLMs}
\label{sec:5}
To clearly articulate the concept of CAC and present an executable example, we implemented a proof of concept using Small Language Models (SLMs). Among Large Language Models, SLMs (typically containing fewer than 10 billion parameters) are increasingly recognized as core components of agentic AI due to their inference efficiency and fine-tuning agility~\cite{belcak2025small}. In CAC, this design choice marks the central trade-off between scalability and explainability. In sufficiently restricted problem spaces, the execution module could instead be implemented with highly deterministic and predictable structures, for example ACT-R-style productions, but such designs become difficult to extend once the problem space expands. Adopting an SLM increases representational flexibility while reintroducing a black-box component. Using fewer parameters therefore does not make the executor fully explainable; it changes how much latent expressive capacity remains inside the model.

\begin{figure}
    \centering
    \includegraphics[width=0.85\linewidth, trim=0cm 6.5cm 13cm 0cm, clip]{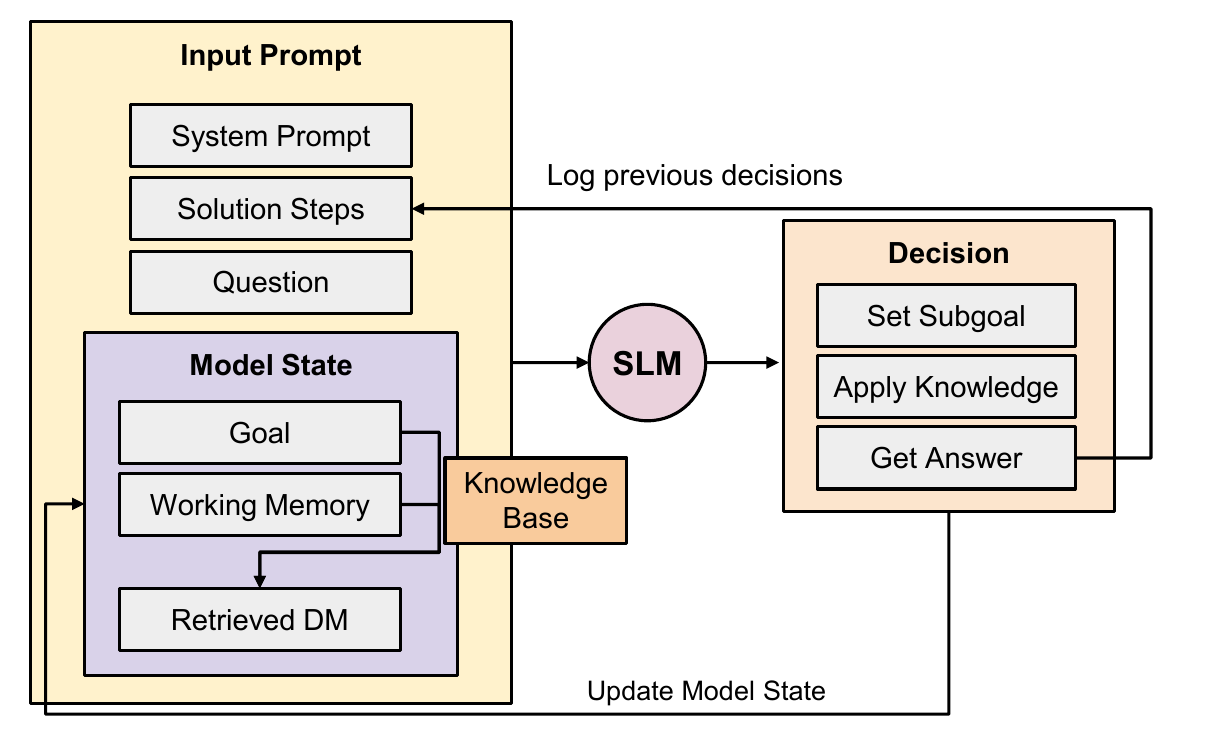}
    \caption{Architecture of the Cognitive Agent. The agent utilizes a Small Language Model (SLM) to process an input prompt containing the current Model State, which is composed of the Goal, Working Memory, and Retrieved Declarative Memory. Operating via a Query-Key-Value mechanism interacting with the Knowledge Base , the SLM makes deterministic decisions to set subgoals, apply retrieved knowledge, or generate a final answer.}
    \label{fig:fig2}
\end{figure}

Inspired by Soar's subgoal formation logic and ACT-R's blending-based declarative memory retrieval system, we implemented a pipeline to train a Cognitive Agent that solves problems by explicitly relying on defined knowledge using an SLM.
\footnote{Experiment code is available at \url{https://github.com/DoniMoon/cognitive_agent_compilation}}
\footnote{A demo containing detailed experiment output samples is available at \url{https://cac-demo.doni.page/}}

\subsection{Implementation of a Prototype Cognitive Language Agent}
We selected the \textit{OLI\_Biology} dataset from the CMU DataShop \footnote{\url{https://pslcdatashop.web.cmu.edu/DatasetInfo?datasetId=1148}} because it satisfies the necessary conditions: a strong dependency on Declarative Memory, the existence of logged student interactions, and publicly available content data. Figure \ref{fig:fig2} depicts the internal operation of the Cognitive Agent during the problem-solving phase, while Figure \ref{fig:fig3} illustrates the overall training loop where the Teacher LLM intervenes in the agent's knowledge by editing the Knowledge Base.

All models in this prototype operate under zero-shot prompting without any fine-tuning. The Cognitive Agent uses Qwen2.5-2B as its core SLM, chosen because it reliably follows the structured output format at the 2B parameter scale. Similarity-based retrieval over the Knowledge Base is performed using embedding-gemma with 300M parameters. The Teacher LLM is gemma-3-27b-it, run locally with 4-bit quantization~\cite{gemma_2025}. Each component receives a system prompt that describes the Cognitive Agent's operational mechanics, the Knowledge Base schema, and the expected output format; full prompt texts are available in the accompanying code repository.\footnotemark[1]

\subsubsection{Defining the Knowledge Schema}
Defining the Knowledge Schema is a critical step tightly coupled with the operational mechanics of the Cognitive Agent. In this implementation, all knowledge is defined using natural language across three specific fields: a ``natural language description of the knowledge,'' the ``Agent's Goal state where the knowledge is applicable,'' and the ``Agent's Working Memory (WM) state where the knowledge is applicable.'' This serves as a simplified abstraction of ACT-R's logic, which retrieves declarative memory through similarity matching with the states of the Goal or Imaginal modules.

\subsubsection{Cognitive Agent}
Figure \ref{fig:fig2} illustrates the architecture of the Cognitive Agent. The interaction between the Model State and the Knowledge Base can be effectively explained through the Query-Key-Value (QKV) mechanism. The agent's Goal and Working Memory are described in natural language, and their computed embeddings function as the \textit{Query}. Every piece of knowledge in the Knowledge Base possesses a ``Goal'' and ``Condition'' detailing its applicability; the embeddings of these fields act as the \textit{Keys}. When the highest-scoring Declarative Memory is retrieved via similarity-based matching, its natural language description becomes the \textit{Value}, which is then injected into the SLM's prompt.

At each step, the SLM selects one of three actions: set a new subgoal, update working memory with retrieved knowledge, or produce a final answer for the current goal. This decision is made by comparing the log-probabilities of the predefined action tokens, so that the action selection itself is deterministic given the prompt. The Goal operates on a hierarchical structure, initially starting from ``solve the question'' and subsequently forming necessary subgoals.

\subsubsection{Teacher LLM}
\begin{figure}
    \centering
    \includegraphics[width=0.85\linewidth, trim=0cm 7.5cm 13cm 0cm, clip]{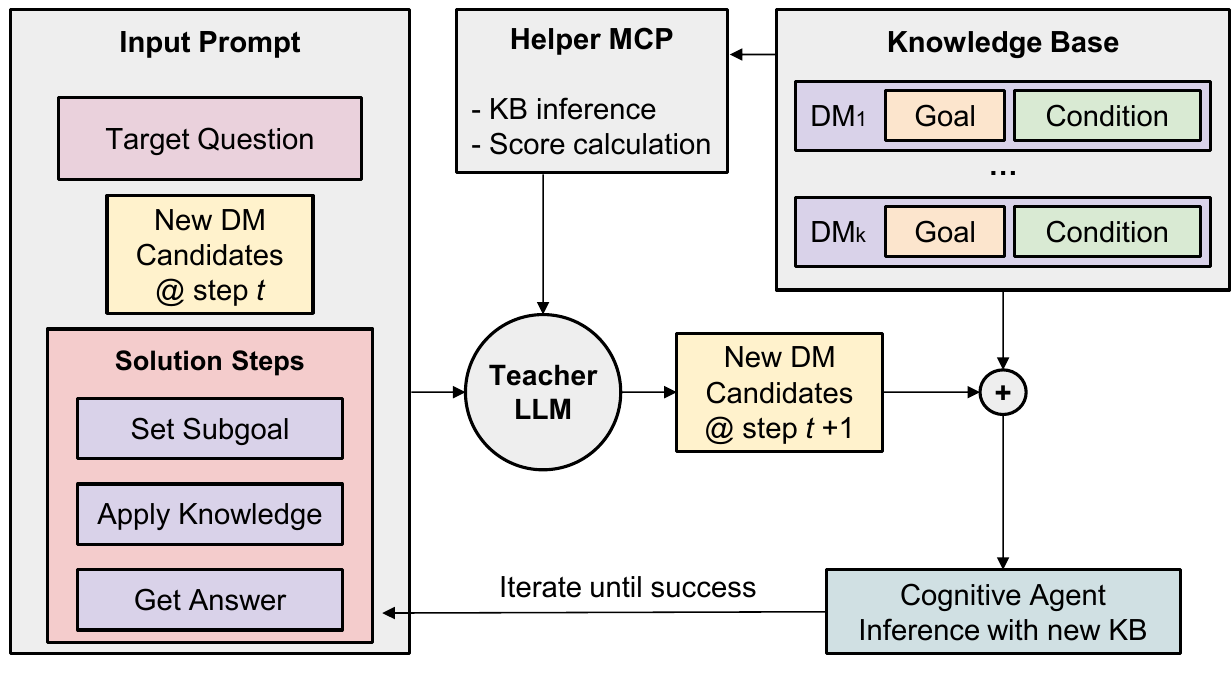}
    \caption{The CAC Training Pipeline. Following a problem-solving attempt, the Teacher LLM analyzes the Cognitive Agent's operational traces and utilizes Helper Model Context Protocols (MCPs) to perform Knowledge Base (KB) inference and similarity score calculations. The Teacher LLM iteratively generates and submits new Declarative Memory (DM) candidates to update the KB until the Cognitive Agent successfully resolves the target question.}
    \label{fig:fig3}
\end{figure}

Figure \ref{fig:fig3} details the operational process of the Teacher LLM. Upon the Cognitive Agent's completion of a problem-solving attempt, the detailed trace, including changes in the Goal and WM, as well as the memories recalled at each step, is recorded and presented to the Teacher LLM. The Teacher LLM receives a prompt that describes the Cognitive Agent's operational mechanism, the current contents of the Knowledge Base, and the failed solution trace. The Teacher can then invoke the following three Model Context Protocol tools:
\begin{enumerate}
 \item A function to assume a hypothetical Model State, execute a retrieval query on the Knowledge Base, and review the top-$k$ results.
 \item A function to calculate the embedding similarity between any two arbitrary sentences.
 \item A function to finalize and submit a list of Declarative Memories (DMs) to be appended to the Knowledge Base.
\end{enumerate}
The Teacher LLM's turn concludes upon invoking the final function. The loop then repeats until the Cognitive Agent reports the correct answer for the given problem. For each problem, the Teacher LLM solidifies a set of DMs to add; thus, as more problems are solved, new DMs accumulate monotonically within the Knowledge Base.
\subsection{Preliminary Findings}


\begin{table*}[t]
\centering
\footnotesize
\setlength{\tabcolsep}{5pt}
\renewcommand{\arraystretch}{1.25}

\caption{A successful Cognitive Agent run. The agent solves a biology MCQ by (i) setting a subgoal, (ii) updating working memory via a retrieved fact, and (iii) answering the subgoal and then the top goal. DM text is shortened for readability; working memory updates are reflected in $\langle R \rangle$ actions.}
\label{tab:cad_trace_example}

\begin{tabular}{
  >{\centering\arraybackslash}p{0.05\textwidth}
  >{\raggedright\arraybackslash}p{0.38\textwidth}
  >{\raggedright\arraybackslash}p{0.45\textwidth}
}
\toprule
\textbf{Step} & \textbf{Retrieved DM} & \textbf{Action} \\
\midrule
0 &
Policy cue: if solving and WM indicates \emph{cellulose is an indigestible plant carbohydrate (dietary fiber)}, proceed toward resolution. &
$[G_0] \rightarrow [G_0, G_1]$ \newline
$\langle G \rangle$ Set subgoal: \emph{Identify which option is indigestible plant-based carbohydrate.} \\
\midrule
1 &
Policy cue: if the current goal is \emph{identify dietary-fiber source} and WM mentions \emph{cellulose is indigestible}, consolidate and move forward. &
$[G_0, G_1] \rightarrow [G_0, G_1]$ \newline
$\langle R \rangle$ Update WM: \emph{Cellulose is indigestible plant-based carbohydrate, matching dietary fiber.} \\
\midrule
2 &
Fact: \emph{Cellulose is dietary fiber in plants and is not digestible by humans.} &
$[G_0, G_1] \rightarrow [G_0]$ \newline
$\langle A \rangle$ Answer subgoal ($G_1$): \emph{Cellulose is indigestible plant-based carbohydrate.} \\
\midrule
3 &
Policy cue: if solving and WM contains \emph{cellulose is indigestible}, answer the MCQ. &
$[G_0] \rightarrow [G_0]$ \newline
$\langle A \rangle$ Answer top goal ($G_0$): \emph{Cellulose is characterized as dietary fiber.} \\
\bottomrule
\end{tabular}

\vspace{0.5em}
\begin{minipage}{0.98\textwidth}
\footnotesize
\textbf{Problem.} ``Which of the following is characterized as dietary fiber?'' Options: A~Glycogen;\enspace \textbf{B~Cellulose};\enspace C~Sucrose;\enspace D~Starch. \\[2pt]
\textbf{Goals.} $G_0$ = Solve the problem.\enspace $G_1$ = Identify which option is indigestible plant-based carbohydrate. \\[2pt]
\textbf{Final scoring.} $p(\mathrm{A})=0.128$,\enspace $\boldsymbol{p(\mathrm{B})=0.845}$,\enspace $p(\mathrm{C})=0.022$,\enspace $p(\mathrm{D})=0.006$.
\end{minipage}
\end{table*}

We executed problem-solving attempts for a total of 27 biology problems.
During training, the agent was iteratively compiled until it produced
the correct answer for each problem, accumulating 188 Declarative
Memory entries with a mean of 3.1 compilation iterations per problem (SD = 2.5). On the 28th problem, the agent failed to produce the correct answer after 150 iterations, and the experiment was terminated.
Table \ref{tab:cad_trace_example} presents a selected example of the
model's operational traces. The following observations highlight
preliminary findings from this proof-of-concept setting:

\textbf{1. Decoupling of Goal and Working Memory Manipulations:} In this prototype, the framework exhibited separable manipulation of goals and working memory under constrained conditions. This suggests the potential to model a specific student failure mode, where a student possesses the requisite knowledge but fails to solve the problem due to difficulty in retrieving the appropriate procedural cue at the correct moment. However, this behavior has not been validated against real student data.

\textbf{2. SLMs as Operational Executors Under Tight Constraints:} The results suggest that an SLM with sufficient performance capacity can execute the minimal, formatted outputs required to drive the Cognitive Agent within tightly constrained action spaces. By enforcing the generation of predefined output formats and leveraging logit-based signals, the system allows certain state transitions to become more inspectable. This should not be interpreted as full explainability; rather, it indicates that a partially latent executor may be constrained to support a degree of explicit control in limited settings.

\textbf{3. Efficacy of Quantized Local Teacher Models:} In this setup, the Teacher LLM was able to generate and inject knowledge in the required format for the Cognitive Agent's operation when guided by a carefully specified prompt. Notably, this was achieved using a 4-bit quantized 27B local model rather than a large commercial API. This observation suggests the possibility that similar pipelines could be implemented with relatively accessible resources, although further validation is required.

\section{Challenges \& Discussions}

Through the proof of concept implementation, several key takeaways and challenges were identified.

\subsection{Curse of Prior Knowledge}
The primary issue stems from the models' inherent prior knowledge. If the model is too large, the SLM answers correctly on its own without needing the proposed cognitive structure; conversely, if the model is too small, its coherence in following rigid instructions breaks down. This is not merely an implementation detail. It defines the core trade-off in CAC between explicit control and scalable coverage. A more deterministic controller would improve predictability, but it would also narrow the problem space the agent can plausibly handle. An SLM extends the reachable problem space, yet it does so by retaining a residual black-box component whose internal prior knowledge cannot be fully inspected. Future work may therefore need alignment methods that suppress internal prior knowledge and force stronger dependence on externally injected knowledge.

\subsection{Fan Effect}
This issue was the primary reason the experiment was terminated at the 28th problem. As the KB grew, similarity-based retrieval increasingly failed to surface newly added DMs, consistent with the fan effect~\cite{anderson1999fan} in which retrieval accuracy degrades as more items share overlapping cues. This caused the agent to stall even after 150 iterations. In several cases the agent retrieved entirely irrelevant knowledge, a phenomenon consistent with deceptive overgeneralization~\cite{an2025deceptive}, where learned retrieval patterns transfer poorly across problem contexts. A more selective retrieval mechanism is needed to mitigate this bottleneck.

\subsection{LLM Dependency}
The framework exhibits a strong dependency on the specific type of LLM utilized. Initially, this proof of concept attempted to identify a knowledge set that would enable multiple SLMs to solve the problems, utilizing six recent instruction-tuned SLMs with under 1 billion parameters. However, because these smaller SLMs failed to follow the structured instructions in varying ways, the experimental design was modified to utilize a single 2B parameter model. A model's suitability as a core component of the Cognitive Agent varies significantly depending on the specific data patterns it was exposed to during pre-training. Using multiple SLM families may increase confidence that the compiled Knowledge Base is not overly model-specific~\cite{moon2022evaluating}, but it does not remove the underlying concern that pretrained models may still contribute knowledge unavailable in the explicit representation.

\subsection{Assessing the Remaining Black-Box Reliance}
The framework therefore requires conceptual tools for judging whether the remaining black-box component is small enough for the intended educational use.

\textbf{1. Knowledge ablation against answer persistence.} One test is to remove candidate core knowledge from the Knowledge Base and verify whether the agent can still answer correctly. If knowledge corresponding to a core concept is removed and the agent still reaches the correct answer, then the remaining SLM should be suspected of relying on prior knowledge rather than on the compiled representation~\cite{moon2022evaluating}.

\textbf{2. Validation through downstream Knowledge Tracing.} A second test is to implement a Knowledge Tracing model using the Knowledge Components generated by the agent and evaluate how well it predicts student errors. If the compiled model underestimates the knowledge required to solve a problem, the resulting Knowledge Tracing model will tend to underpredict incorrect responses. A remaining methodological challenge is to enforce \textit{knowledge-performance monotonicity}: higher inferred mastery of the required knowledge components should monotonically increase the predicted probability of correctness.

\subsection{Collaborative Cheating}
Instances were identified where the Teacher LLM directly provided the final answer to the student agent, bypassing the intended cognitive process. This issue can be mitigated using the methodology discussed in Section \ref{sec:goodkc}. Because knowledge that explicitly dictates a specific answer lacks generalizability to other similar problems, such instances can be programmatically identified and filtered out during the compilation process.

\section{Conclusion}

This paper introduced Cognitive Agent Compilation (CAC), a framework for compiling problem-solving knowledge extracted from a strong Teacher LLM into explicit structures inside a target Cognitive Agent. The current contribution is limited to a proof of concept showing that goal formulation, working memory updates, and declarative retrieval can be partially separated and inspected within a constrained pipeline. The study does not yet establish CAC as a validated learner model or a complete solution for Knowledge Tracing. Instead, it identifies the main design tensions that must be resolved, especially the trade-off between scalable black-box executors and fully explicit controllers, the persistence of prior knowledge, and retrieval failures as the Knowledge Base grows. Progress on this agenda will require methods for testing whether success truly depends on compiled knowledge~\cite{moon2022evaluating} and evaluation against student data, including the preservation of knowledge-performance monotonicity in downstream Knowledge Tracing models. 

\bibliographystyle{splncs04}
\bibliography{mybibliography}

\end{document}